\pdfoutput=1

\documentclass[11pt]{article}

\usepackage{emnlp2023}

\usepackage{times}
\usepackage{latexsym}
\usepackage{subcaption}
\usepackage{graphicx}
\usepackage[T1]{fontenc}

\usepackage[utf8]{inputenc}

\usepackage{microtype}

\usepackage{inconsolata}

\title{Large Language Models are legal but they are not: Making the case for a powerful LegalLLM }

\author{Thanmay Jayakumar \qquad  Fauzan Farooqui$^*$ \qquad Luqman Farooqui$^*$ \\
  Visvesvaraya National Institute of Technology,
  Nagpur, India \\
  \texttt{\{thanmayjayakumar, fauzanfarooqui7,  luqmanfarooqui99\}@gmail.com}
}

\begin{document}
\maketitle
\begin{abstract}
Realizing the recent advances in Natural Language Processing (NLP) to the legal sector poses challenging problems such as extremely long sequence lengths, specialized vocabulary that is usually only understood by legal professionals, and high amounts of data imbalance. The recent surge of Large Language Models (LLMs) has begun to provide new opportunities to apply NLP in the legal domain due to their ability to handle lengthy, complex sequences. Moreover, the emergence of domain-specific LLMs has displayed extremely promising results on various tasks. In this study, we aim to quantify how general LLMs perform in comparison to legal-domain models (be it an LLM or otherwise). Specifically, we compare the zero-shot performance of three general-purpose LLMs (ChatGPT-20b, LLaMA-2-70b, and Falcon-180b) on the LEDGAR subset of the LexGLUE benchmark for contract provision classification. Although the LLMs were not explicitly trained on legal data, we observe that they are still able to classify the theme correctly in most cases. However, we find that their mic-F1/mac-F1 performance is up to 19.2/26.8\% lesser than smaller models fine-tuned on the legal domain, thus underscoring the need for more powerful legal-domain LLMs.

\end{abstract}
\def\thefootnote{*}\footnotetext{These authors contributed equally}\def\thefootnote{\arabic{footnote}}

\section{Introduction}
\label{sec:intro}
 Legal professionals typically deal with large amounts of textual information on a daily basis to make well-informed decisions in their practice. This can become very tedious and demanding due to the overwhelming amount of data they must manage and the meticulous attention to detail necessary to maintain the required precision in their work. Thanks to the rise of LLMs, many tasks such as sentiment analysis, named entity recognition, information retrieval, etc. can now be handled by neural models. Though this holds true for the legal domain as well \cite{sun2023short}, they aren't used to make direct decisions. Nevertheless, these automated systems that produce legal predictions and generations, are predominantly useful as advisory tools for legal practitioners that can augment their decision-making process.

Transformers \cite{vaswani2017attention} have become the de facto method for many text classification and multiple choice question answering tasks. BERT \cite{devlin-etal-2019-bert}, a transformer-encoder, and its derived models like RoBERTa \cite{roberta} are commonly employed in legal NLP tasks. Pre-training such models on legal corpora can help a model adapt to a specific domain by fine-tuning it with domain-specific data. LegalBERT \cite{chalkidis-etal-2020-legal} is one such BERT model that was trained on legal-oriented data. CaseLawBERT \cite{caselaw}, PoL-BERT \cite{henderson2022pile}, and LexLM \cite{chalkidis-etal-2023-lexfiles} are a few more BERT-based variants pre-trained for the legal domain. Although they show remarkable performance on various legal tasks in comparison with general-purpose BERT models, one limit of these models is that BERT's input size can only incorporate a maximum of 512 tokens. For short sequences this may seem enough, but in the case of long documents commonly found in the legal domain, where input texts can go over 5000 tokens (and requiring \textit{even} more in few-shot settings), it can be a severe drawback as a lot of important information will get truncated.

Due to this limit, BERT-based models aren't employed as-is in long-document tasks. Typically, methods like hierarchical attention are utilized where the long document is split into segments of max length (512 in the case of BERT models) and these segments are independently encoded. These segment embeddings are then aggregated with stacked transformers to get the overall encoding of the entire document. Similarly, recurrent transformers \cite{dai-etal-2019-transformer, yang2019xlnet,ding-etal-2021-ernie} were proposed to process long documents by encoding its representation from individual segments in a recurrent fashion. Sparse attention is another method that has been proposed to tackle long sequence inputs \cite{ainslie-etal-2020-etc, zaheer2020big}. Longformer \cite{beltagy2020longformer} uses a combination of local and global attention mechanisms to save on computational complexity and enables the processing of up to 4096 tokens. A number of other works \cite{dai-etal-2022-revisiting, mamakas-etal-2022-processing} show that transformer-based architectures that can capture longer text boast major benefits, even more so when augmented with strategies like sparse-attention and hierarchical networks. This again underlines an important direction for verbose legal datasets.

Our contributions can be summarized as follows:
\begin{itemize}

    \item We conduct experiments to compare and analyze the zero-shot performance of three general LLMs to that of start-of-the-art in-domain models on the LEDGAR subset of LexGLUE \cite{chalkidis-etal-2022-lexglue}. We analyze our results, quantify whether LLMs conform to expected advantages, and provide insights for further research.
    \item We provide an overview of the most recent LLM research, the benchmarks and datasets developed for legal NLP, the challenges faced when applying them to legal tasks, and popular approaches that solve them. We believe this to be a useful primer for anyone looking to get a bird's eye view of the field.
\end{itemize}

\section{Related Work}
\label{sec:related}
In this section, we outline the relevant research on LLMs, efforts in using them for legal domain tasks, and finally the benchmarks and datasets.

\subsection{Large Language Models}
\label{subsec:llm}
\textbf{OpenAI GPT:} GPT (Generative Pre-trained Transformer) \cite{Radford2019LanguageMA, brown2020language} and the popular ChatGPT variant developed by OpenAI are a family of large-scale proprietary transformer-decoder models pretrained to perform generative and language modeling tasks, and allow a reasonable context length sufficient to carry out long-document processing. For instance, GPT 3.5 supports a maximum of 4096 tokens, and GPT 4 allows a stunning maximum of 32,768, ideal for data consisting of long sequences.\\

\noindent
\textbf{Google PaLM}: PaLM (Pathways Language Model) \cite{chowdhery2022palm, anil2023palm} is a proprietary LLM having 540 billion parameters that was trained on the Pathways architecture. Although PaLM was initially trained to handle sequence lengths of up to 2048 tokens, it was increased to 8096 in the 340 billion parameter PaLM 2 for a longer comprehension of the input.\\

\noindent
\textbf{BigScience BLOOM}: BLOOM (BigScience Large Open-science Open-access Multilingual Language Model) \cite{bloom} is a group of open-source multilingual LLMs, the largest having 176 billion parameters. It encompasses 46 natural and 13 programming languages, facilitating sequence lengths of up to 2048 tokens.\\

\noindent
\textbf{Meta LLaMA:} LLaMA (Large Language Model Meta AI) \cite{touvron2023llama} is a collection of open-source foundation language models ranging from 7 billion to 70 billion parameters. It was pre-trained natively on 2048 input tokens, but recent research has shown that the context length of LLMs can be extended efficiently with minimal training steps \cite{yarn}, leading to a release of two variations of LLaMA that have an astounding context length of 64k and 128k.\\

\noindent
\textbf{TII Falcon\footnote{\url{https://falconllm.tii.ae/falcon.html}}}: This work by the Technology Innovation Institute (TII) boasts of being the largest open-source model to date, while also ranking highest on the HuggingFace Leaderboard.  It includes models with 180B, 40B, 7.5B, and 1.3B parameters (context window size of 2048) trained on TII’s RefinedWeb dataset \cite{refinedweb}.

\subsection{LLMs on the legal domain}
\label{subsec:lllm}

\textbf{LexGPT: }\cite{LexGPT} finetune GPT-J models on the Pile of Law dataset \cite{henderson2022pile}, and is the best-performing LLM that has been finetuned for legal use cases (LegalLLM) at the time of writing. They experiment with generative models for legal classification tasks and observe that fine-tuning such out-of-the-box GPTs still results in low performance when compared to discriminative models. This insightfully shows the need to bridge the gap between powerful LLMs and the legal domain.

\noindent
\textbf{PolicyGPT: } \cite{policygpt} demonstrate that many LLMs in zero-shot settings perform remarkably well when tasked with text classification of privacy policies. Though specific, this shows how a LegalLLM may hold promise in enhancing performance on other general tasks.\\

\noindent
\textbf{Zero-and-Few-shot GPT: }\cite{chalkidis2023chatgpt} conduct experiments most similar to our study. They evaluate the performance of ChatGPT on the LexGLUE benchmark in both zero-shot and few-shot settings (for the latter, examples were given in the instruction prompt, which seems to benefit the model when the number of examples and labels are around the same). They find that ChatGPT performs very well, but severely lacks in performance compared to smaller models trained on in-domain datasets.

Resonating with these findings, the work of \cite{savelka2023unlocking} investigates how an LLM (a GPT model) performs on a semantic annotation task in zero-shot settings, without being fine-tuned on legal-domain datasets. The LLM is primed with a short sentence description of each annotation label and is tasked with labeling a short span of text. They observe that while the LLM performs surprisingly well given the zero-shot setting, its performance was still far off from the model that was trained on the in-domain data. In summary, both studies highlight the potential fine-tuned LLMs can bring to the legal domain.

\subsection{Datasets and Benchmarks}
\label{subsec:benchmarks}
\textbf{LexGLUE: } \cite{chalkidis-etal-2022-lexglue} present a unified evaluation framework for legal tasks to benchmark models. The datasets and tasks were curated from other sources of data considering various factors such as availability, size, difficulty, etc. They present scores for various Pre-trained Language Models (PLMs) on their benchmark. They point out interesting results that suggest that PLMs fine-tuned on general legal datasets and tasks do perform better, albeit PLMs fine-tuned on only one sub-domain don't improve on performance on the same sub-domain. Put together, their observations point out the need for a general LegalLLM (powerful enough to outperform other models on all criteria of the benchmark).vspace{-4pt}\\\\\
\textbf{LegalBench: } \hspace{-30pt} \cite{legalbench} This benchmark comprises 162 tasks representing six distinct forms of legal reasoning and outlines an empirical evaluation of 20 LLMs. They demonstrate how LegalBench supports easing communication between legal professionals and LLM developers by using the IRAC framework in the case of American law. They observe that LLMs typically perform better on classification tasks than application-based ones. They also find that for some tasks, in-context examples are not required, or only marginally improve performance. They thus conclude that the task performance in LLMs is mostly driven by the task description used in the prompt.\\\\
\textbf{Pile of Law: } \hspace{-25pt} \cite{henderson2022pile} The surge in LLM development emphasizes the need for responsible practices in filtering out biased, explicit, copyrighted, and confidential content during pre-training. Present methodologies are ad hoc and do not account for context. To address this, Pile of Law, a growing 256GB dataset of open-source English legal and administrative data, was introduced to aid in legal tasks. This paper outlines a method for filtering legal-domain text while handling associated trade-offs. It aids in understanding government-established content filtering guidelines and illustrates various ways to learn responsible data filtering from the law. \\\\
\textbf{MultiLegalPile: } \hspace{-25pt}\cite{chalkidis-etal-2021-multieurlex} The MultiLegalPile is a 689 GB substantial dataset that spans 24 EU languages across 17 jurisdictions. It addresses the scarce availability of multilingual pre-training data in the legal domain, encompassing diverse legal data sources with varying licenses. 
In certain languages, monolingual models substantially outperform the base model, achieving language-specific SotA in five languages. 
In LexGLUE, English models secure SotA in five of seven tasks.

\begin{figure*}[ht]
\centering
\begin{subfigure}{0.5\textwidth}
  \centering
  \includegraphics[width=\linewidth]{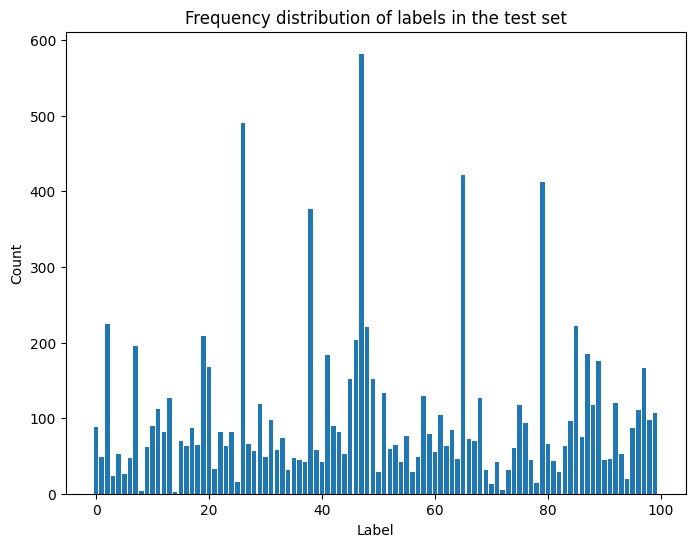}
  \label{fig:sub1}
\end{subfigure}%
\begin{subfigure}{0.5\textwidth}
  \centering
  \includegraphics[width=\linewidth]{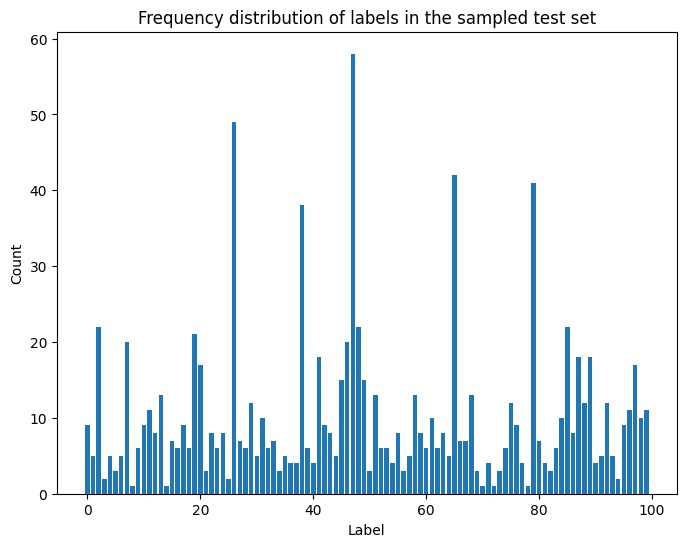}
  \label{fig:sub2}
\end{subfigure}
\caption{The frequency distributions of the 100 LEDGAR labels\\in the original LEDGAR test set from LexGLUE (left); and in our sampled test set of 1000 examples (right)}
\label{fig:ledgar-dist}
\end{figure*}

\section{Experimental Setup and Results}
In this section, we describe our experimental approach, along with specifics of our evaluations.
\label{sec:experiments}

\subsection{Dataset and Metrics}
We use the LEDGAR \cite{tuggener-etal-2020-ledgar} subset of the LexGLUE benchmark for our experiments due to its readiness to work on LLMs (for example, the other datasets have label indices alone, not the actual label names). The dataset was loaded through the HuggingFace Datasets library \cite{datasets}. In this benchmark, given a provision contract, the model is tasked with classifying the contract into one of 100 EDGAR theme labels. As mentioned, there is a high imbalance of data in datasets containing legal corpora. Figure \ref{fig:ledgar-dist} shows the label distribution in the LEDGAR subset benchmark. This could result in difficulties such as biased models and poor classification scores. To better report model evaluations in such settings, the F1-score is usually reported instead of accuracy. 
Moreover, both macro-F1 and micro-F1 scores are usually reported. For imbalanced datasets, the former more accurately reflects the classifier's performance as the latter skews the metric towards the larger-proportion datasets, which is why the macro-F1 scores are typically lower than the micro-F1 ones in these scenarios.

As for the sequence lengths, \cite{chalkidis2023chatgpt} report the average token length of the instruction-following examples in all the LexGLUE subsets - the highest being 3.6k tokens. This restricts the capability of LLM performance due to truncation as noted earlier, and this is also highlighted in their study: few-shot settings could not be evaluated for datasets having an average token length of more than 2k for a single example, and in many cases, the prompts were already truncated up to 4k tokens (the maximum limit of ChatGPT). The average token length of the LEDGAR subset is 0.6k. 

\subsection{Setup}
As baselines, we take three LLMs: ChatGPT (20b), LLaMA-2 (70b), and Falcon (180b). Since the models are very large, we use HuggingFace Chat for LLaMA and Falcon. Due to this constraint, we only evaluated on a subset of 1,000 examples. However, we made sure that the subset had a label frequency distribution close to the original dataset (Refer Figure \ref{fig:ledgar-dist}) so that the evaluations remain generalizable.

We use zero-shot prompting to evaluate the above-mentioned LLMs, building on the benefits as explained earlier by other works \cite{policygpt, legalbench}. Further, in the custom instructions (ChatGPT) and system instructions (HuggingChat), we enter the list of EDGAR theme classes that the model should choose from. In the same fashion, to ensure that the model does not generate anything out of the list, we explicitly mention this constraint as an instruction. The exact instructions that we use are provided in the appendix.

\subsection{Results and Discussion}
For our experiments, we use three baseline general-purpose chat variants - ChatGPT (20b), Falcon (180b), and LLaMA-2 (70b) - and present the results in Table \ref{tab:results}. General-purpose LLMs perform worse than smaller in-domain models. The best general LLM, Falcon-Chat, performs 19.2\% mic-F1 and 26.8\% mac-F1 lower than the best in-domain model, LegalBERT, which itself is much smaller than the LexGPT, the current LegalLLM. Our findings echo that of \cite{chalkidis2023chatgpt}. 

\begin{table}[h]
\centering
\begin{tabular}{|l|l|l|l|}
\hline
\textbf{Model       }& \textbf{mic. F1}       & \textbf{mac. F1 }& \textbf{\# params.} \\ \hline 
Falcon-Chat & 70.9          & 60.7  & 180b\\ \hline
LLaMA-Chat  & 70.4          & 59.6  & 70b \\ \hline
ChatGPT     & 70.6          & 58.7  & 20b\\ \hline
LexGPT             & \textit{83.9}    & \textit{74.0} & 6b \\ \hline
LegalBERT          & \textbf{88.2} & \textbf{83.0}  & 0.11b \\ \hline
\end{tabular}
\caption{Comparison of general LLMs (first three models, tested on a zero-shot setting by us) to models fine-tuned on legal-domain datasets (last two). The current LegalLLM is LexGPT, but the much smaller LegalBERT shows state-of-the-art performance on LEDGAR.}
\label{tab:results}
\end{table}

Notably, for class labels with only one example in our sampled test set, the three chat-variants surprisingly show the same results: they fail to predict them correctly, except the \textit{Qualification} label (the others being \textit{Assigns, Books, Powers} and \textit{Sanctions}. Similarly, \textit{Indemnity} is always misclassified as \textit{Indemnifications} (three examples in total). Further, labels that are semantically similar are frequently mislabeled by the models (like \textit{Indemnity} and \textit{Indemnifications} as pointed out earlier). For example, (\textit{Taxes}, \textit{Tax Withholdings} and \textit{Withholdings}) is almost always labeled as \textit{Tax Withholdings} by all the models. (\textit{Jurisdictions, Submission To Jurisdiction, Consent To Jurisdiction}) is almost always labeled as \textit{Submission To Jurisdiction} in the case of ChatGPT and \textit{Jurisdiction} in the case of Falcon and LLaMA. As for (\textit{Applicable Laws, Governing Laws, Compliance With Laws}), we observe that \textit{Governing Laws} was easiest to predict with an average accuracy of 90\%, and \textit{Compliance With Laws} with 80\%, but \textit{Applicable Laws} performs very poorly with 0\% accuracy for LLaMA and Falcon and 20\% for ChatGPT - predicting only one from a total of 5 samples correctly. However, in the case of (\textit{Payments, Fees, Interests}), the models seem to predict them correctly in about 60\% of the cases, with \textit{Payments} appearing at least once for \textit{Fees} and \textit{Interests}. On average, only 95 of the 100 classes in the reference labels are present in the predictions.

\subsection{Subjective Analysis}
Our findings highlight that the perceived advantages LLMs have over BERT-based models (such as the sheer amount of large parameters, extended context length, and the amount of pre-training knowledge), cannot substitute for the obvious edge in-domain data gives to the much smaller models. Even when the LLM is trained so (LexGPT), it couldn't perform as well as the discriminative model (LegalBERT). This could be expected as the latter is more naturally suited for the benchmark's classification tasks than generative models which are prone to issues like hallucination. Our label-wise findings reflect this too.

However, the current legal benchmarks are limited to NLU tasks. In general, it would be ideal to have a powerful LegalLLM that can perform both generative and discriminative tasks. Our findings show that there is a unique challenge in the legal domain: if we have to build a better LegalLLM, we need to find better methods to take advantage of the in-domain legal data for LLMs as simply fine-tuning it doesn't seem to be enough. As the authors of LexGPT mention, reinforcement learning from human feedback could be extremely helpful in improving LexGPT, providing ways for the first LegalLLM to produce state-of-the-art results. 

However, if we limit the application of legal models to NLU tasks, our findings turn optimistic. The results show that the LLMs' ability to process large context may not be necessary for classification - we hypothesize this could be because verbose legal text could turn out to have very similar semantic content, so the additional context may not be as useful as expected. This hypothesis could be echoed by findings from \cite{shaikh2020predicting}, who show that a careful selection of a handful of textual features in a verbose dataset is strong enough to help statistical models achieve high accuracies for binary classification.

This in fact should be good news for NLU, as it means legal practitioners can avoid having to use or train unnecessarily large or expensive models (both carbon-wise and cost-wise). Much smaller in-domain models like LegalBERT are nevertheless superior and should be used for practical applications, as suggested by \cite{chalkidis2023chatgpt}

\section{Conclusion}
In this work, we examine three general-purpose LLMs' zero-shot performance on a multi-class contract provision classification task using the LEDGAR dataset of LexGLUE. Our study shows that these LLMs, even though not explicitly trained on legal data, can still demonstrate respectable theme classification performance but are easily overshadowed by smaller in-domain models. The results highlight the need for better LegalLLMs. In light of this, we also present a review of related datasets and models, which we hope will help get an overview of the field.

\bibliographystyle{acl_natbib}
\bibliography{anthology,custom}

\appendix

\section{Custom Prompt}
For reproducibility, we present the prompts that we use for all our experiments. The following is the entry to the Custom Instructions setting of ChatGPT. For HuggingChat, we simply provide both the instructions to the Custom System Prompt box.\\\\
\textit{\textbf{What would you like ChatGPT to know about you to provide better responses?} I want you to be an EDGAR contract provision classifier. Given a contract provision, you should correctly identify the EDGAR theme. Do not give any explanations.}\\\\
\textit{\textbf{How would you like ChatGPT to respond?} One answer from the following list: [ \{\{paste the list here\}\} ]. Do not give an option that is not in the list.}

\end{document}